\documentclass[lettersize,journal]{IEEEtran}
\usepackage{amsmath,amsfonts}
\usepackage{algorithmic}
\usepackage{algorithm}
\usepackage{array}
\usepackage[caption=false,font=normalsize,labelfont=sf,textfont=sf]{subfig}
\usepackage{textcomp}
\usepackage{stfloats}
\usepackage{url}
\usepackage{verbatim}
\usepackage{graphicx}
\usepackage{cite}
\usepackage{colortbl}
\usepackage{color}
\usepackage{cuted}
\usepackage{caption}
\usepackage{booktabs}
\usepackage{multirow}
\usepackage{pifont} 
\usepackage[table,xcdraw]{xcolor}
\begin{document}

\title{Can Vision–Language Models Truly Understand Images? A Benchmark Beyond Commonsense Bias}

\author{Ling~Lin, Heng~Su, Xiangzeng~Liu, Yang~Bai,  Congcong~Zhu, Jingrun~Chen

\IEEEcompsocitemizethanks{

\IEEEcompsocthanksitem

}}

\markboth{}%
{Shell \MakeLowercase{\textit{et al.}}: A Sample Article Using IEEEtran.cls for IEEE Journals}


\maketitle

\begin{abstract}
Multimodal large language models (MLLMs) have achieved impressive performance on a wide range of vision–language tasks. However, it remains unclear whether their predictions are grounded in the visual content itself or are largely driven by statistical co-occurrence patterns and commonsense correlations learned from large-scale image–text corpora. Existing vision–language benchmarks predominantly consist of samples that conform to human commonsense, allowing models to achieve strong performance by exploiting dataset priors rather than genuinely understanding image content. To address this limitation, we present ContraSense, a systematically constructed benchmark for evaluating faithful vision–language understanding. ContraSense consists of carefully curated counter-intuitive image–text pairs that deliberately violate common expectations, substantially reducing the effectiveness of statistical biases and requiring models to rely on visual evidence for reasoning. We conduct comprehensive evaluations on a diverse set of representative open- and closed-source MLLMs. The results reveal a substantial performance gap between current MLLMs and humans: even the strongest closed-source model achieves only about 60\% accuracy, compared with 96.04\% for human participants. These findings demonstrate that current MLLMs still struggle with faithful visual understanding and highlight the need for more robust multimodal reasoning grounded in image content. The ContraSense benchmark is publicly available at \url{https://huggingface.co/datasets/LingLin00/ContraSense}.

\end{abstract}


\begin{figure*}
  \centering
  \includegraphics[width=\textwidth]{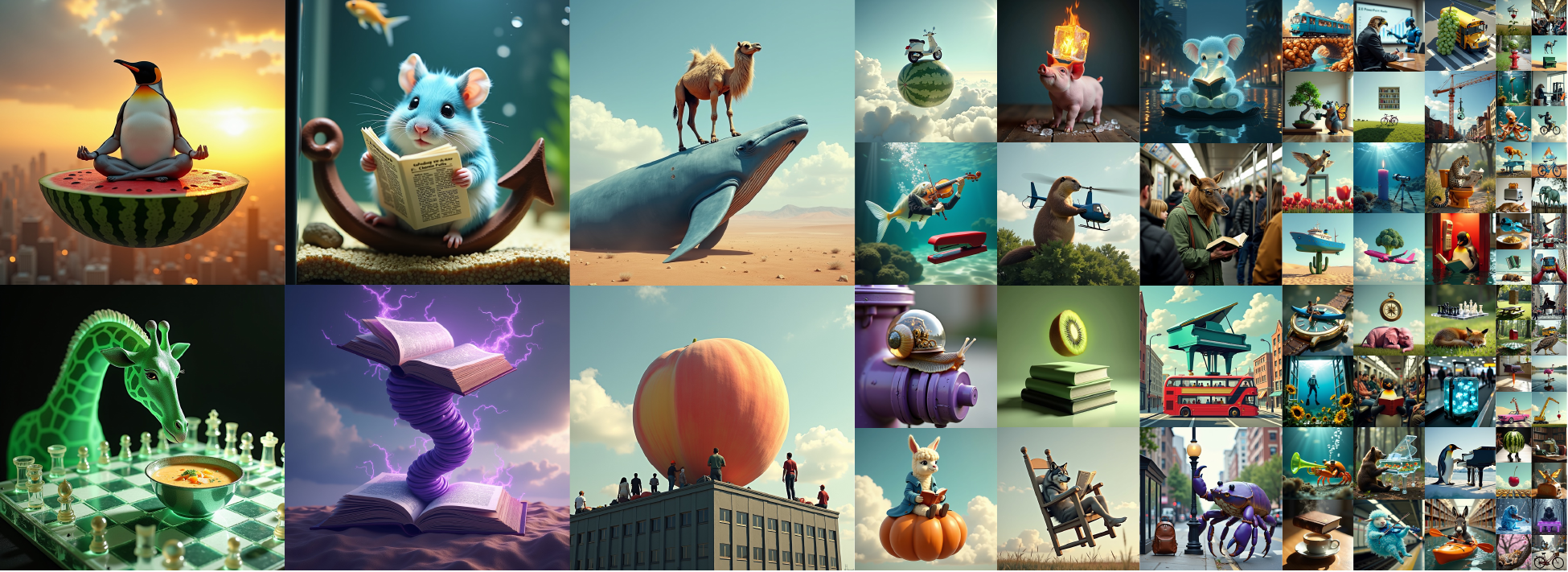}
  \captionof{figure}{
    We propose a new benchmark called ContraSense, which is both creative and imaginative. The benchmark consists of 4,698 carefully crafted and manually calibrated high-resolution (1024 $\times$ 1024 pixels) text QA pairs. Each image contains at least one counterintuitive scenario designed to assess whether visual-language models can provide accurate responses based on the image content itself without relying on common-sense prior knowledge from the training data. This benchmark provides a unique and powerful tool for studying the robustness and real-world visual understanding capabilities of VLMs.
  }
  \label{fig: contrasense examples}
  \vspace{-0.5em}
\end{figure*}

\section{Introduction}
\label{introduction}
\IEEEPARstart{W}{ith} the rapid development of Vision-Language Models (VLMs), related models have made significant progress in a variety of tasks such as image-text alignment, Visual Question Answering (VQA), and Image Captioning~\cite{chen2022pali,alayrac2022flamingo,kosmos22023,liu2023llava,gpt4o,team2024gemini,driess2023palme,shi2025multi}. They have demonstrated excellent cross-modal migration and generalization capabilities and are able to describe image content more accurately or provide seemingly reasonable answers to questions posed by images~\cite{gpt4o, wang2024headdiff,team2024gemini}. Yet this prompts a fundamental question: Do these models actually \emph{see} images? In other words, are their outputs truly grounded in the visual content, or do they instead lean on linguistic/visual \emph{a priori} knowledge—“common sense’’ built from frequent statistical co-occurrences~\footnote{Statistical co-occurrence refers to potential associative knowledge that occurs frequently in the training data. For example, in multimodal data, ``fish'' often co-occurs with ``water'', which causes the model to be more inclined to judge the environment as ``water'' when identifying fish, even if the actual scene is not. In this paper, we call this kind of potential co-occurrence knowledge ``common sense''.} in the multimodal training data? Such co-occurrences can yield superficially correct responses based on high-frequency patterns, allowing models to score artificially high without deep image analysis~\cite{raj2024biasdora}.
This phenomenon can trigger serious consequences in real-world applications, especially when models are deployed in scenarios where accurate perception and reasoning are highly required. For example, in autonomous driving, models may ignore the actual state of lights due to statistically common patterns such as ``red-stop'' or ``green-forward'' and thus make incorrect judgments in abnormal lighting or occlusion situations.

An intuitive idea for evaluation is to directly test the multimodal model's ability to understand image content using existing datasets. However, existing datasets generally suffer from statistical co-occurrence during the construction process. For example, images of camels obtained through natural collection methods often appear in a desert background. Therefore, when the model is asked ``where do camels live'', even without understanding the content of the image, it can give a seemingly correct answer of ``desert'' based on the statistical co-occurrence pattern learned from training. This coupling effect makes it difficult to judge whether the correct answer of the model is based on the real understanding of the semantics of the image or only on the co-occurrence patterns learned from the training corpus. Therefore, although existing datasets play an important role in multimodal model evaluation, their applicability may be limited in specific contexts, especially when decoupled statistical co-occurrence is required to evaluate the model's real image comprehension. This strongly motivates us to construct a new evaluation benchmark that should deliberately break the statistical co-occurrence relationship of the training data so that the model cannot obtain the correct answer through a priori knowledge or frequent co-occurrence but has to rely on a deep understanding of the image content itself. By minimizing the impact of statistical co-occurrences on the answers, the benchmark can more accurately measure model's ability to understand the image semantics truly.

In terms of dataset construction, the existing approach of collecting natural image data often inevitably introduces statistical co-occurrence bias, which makes it difficult to assess the realistic comprehension of image content by multimodal models in the inference process. To alleviate the problem, we propose a generation-based approach to construct a multimodal benchmark, \textbf{ContraSense Benchmark}, for counterintuitive reasoning. Specifically, we crafted a set of prompts that cover five types of common knowledge violations, including \textit{contextual violation, attribute violation, size/proportion anomalies, physical impossibility,} and \textit{function/state perversity} During the data generation process, we randomly select at least one counterintuitive type and construct a conflict situation between the two categories to guide the Large Language Model (LLM) in the form of a prompt to generate the corresponding counterintuitive image captions as well as six question-answer pairs (QA pairs) in a regularized format. Subsequently, we transformed the generated captions into high-resolution (1024 $\times$ 1024) images using a text-to-image model. To ensure the consistency of the images and the quality of the questions and answers, we invited five human experts to manually proofread the image descriptions and the corresponding question-answer pairs and correct the formatting, as well as eliminate samples that are not consistent with the images and text, or that can be answered by logical reasoning. In the end, we constructed the ContraSense Benchmark with 4,698 high-quality counterintuitive QA samples covering a wide range of scenes and styles, including indoor and outdoor environments, painting styles, 3D rendering styles, \textit{etc}. Fig.~\ref{fig: contrasense examples} shows some counterintuitive images from the ContraSense dataset.

In order to comprehensively assess the ability of visual language models (VLMs) to understand image content realistically, we further propose the Contra-Acc metric, which measures the performance of the models in processing counterintuitive images from multiple dimensions. 
The experimental results show an average performance degradation of about 37\% for these models on the ContraSense Benchmark compared to regular image tasks, reflecting the significant performance gap of current VLMs in counterintuitive scenarios.

The main contributions are summarized as follows:
\begin{itemize}
    \item [$\bullet$] We have carefully constructed ContraSense, a image-text multimodal dataset dedicated to counterintuitive comprehension, which contains a total of 4,698 well-designed counterintuitive QA examples, aiming to systematically examine whether a multimodal large model can faithfully comprehend and align image content in the face of common-sense biases. 
    \item [$\bullet$] We propose a new evaluation metric, called Contra-Acc metric, to more comprehensively measure a model's ability to understand an image's semantics truly. The metric is measured in six counterintuitive dimensions, aiming to exclude common sense from interfering with the model's behavior and, thus, more accurately measure the model's fidelity to the content of the image itself.
    \item [$\bullet$] We conducted experimental research on several mainstream multimodal large models. The results show that existing models generally suffer from significant performance bottlenecks in counterintuitive tasks, and they still suffer from significant robustness deficiencies in handling scenarios that violate common sense.
\end{itemize}

\section{Related Work}
\label{related work}

\noindent\textbf{Comprehensive Benchmarks for Large VLMs.} 
A series of large-scale, general-purpose benchmarks~\cite{wang2024toward,li2025ustc,li2023common,wang2024alignment,hou2025domainverse} have emerged to evaluate the capabilities of multimodal large language models (MLLMs).
SEED-Bench and its hierarchical successor SEED-Bench-2 provide 19K and 24K multiple-choice questions respectively, spanning up to 27 evaluation dimensions across image, video, and generation tasks~\cite{seedbench2023,seedbench2_2023}.
MMBench introduces a 3K bilingual test set with automatic grading, enabling direct comparisons between English and Chinese models~\cite{mmbench2023}.
MME categorizes 14 subtasks into "perception" and "cognition" branches for finer-grained diagnostic insights~\cite{mme2023}, while MM-Vet emphasizes integrative skills such as optical character recognition (OCR), spatial reasoning, and mathematical understanding~\cite{mmvet2023}.
Beyond static leaderboards, LVLM-eHub aggregates over 40 public datasets into an online evaluation hub, providing continuously updated model rankings~\cite{lvlmehub2023}.
Traditional vision-reasoning corpora, such as GQA, remain popular for compositional question answering and are commonly included in benchmark suites~\cite{gqa2019}.
Additionally, specialized diagnostic benchmarks such as POPE, which probes models for object hallucinations, are now frequently reported alongside holistic metrics to highlight specific failure modes~\cite{pope2023}. Zhou et al.~\cite{zhou2023rome} introduce a novel probing dataset named ROME to evaluate whether the SOTA pre-trained vision-language models have the reasoning capability to correctly interpret counter-intuitive content. Bitton-Guetta et al.~\cite{bitton2023breaking} released the WHOOPS! dataset and benchmark to evaluate models' performance in visual common-sense understanding. This dataset consists of images deliberately constructed by designers using publicly available image generation tools (such as Midjourney) to violate common sense.

However, despite their comprehensive coverage, these benchmark suites inherently reflect commonsense regularities prevalent in natural images (e.g., \textit{camel $\leftrightarrow$ desert}, \textit{plane $\leftrightarrow$ sky}), or it does not primarily emphasize the impact of disrupting statistical co-occurrence on model conclusions to achieve the goal of verifying whether the model remains faithful to the specific content of the image, allowing shortcut-seeking models to answer correctly by exploiting frequent co-occurrence patterns rather than genuine visual understanding. Furthermore, multiple-choice or short-answer formats predominant in these benchmarks favor superficial keyword matching, rarely confronting models with multifaceted contradictions involving context, physics, and function. Unlike the aforementioned work, ContraSense's core objective is to construct a class of problem samples for which answers cannot be directly inferred from statistical co-occurrence patterns. This tests whether models genuinely possess the ability to understand and reason about image semantics. To achieve this, we systematically designed counterintuitive scenarios, ensuring both task design and problem construction revolved strictly around this central principle. Consequently, existing benchmarks inadequately assess how VLMs behave when commonsense priors conflict with visual evidence—a critical evaluation gap explicitly addressed by our proposed ContraSense benchmark.

\section{The ContraSense}
\begin{figure}[t]
  \centering

   \includegraphics[width=0.55\linewidth]{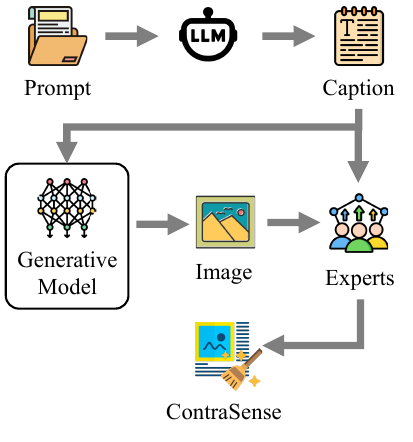}

   \caption{Pipeline of counterintutive data collection.}
   \label{fig: pipeline}
\end{figure}

To achieve the goal of constructing counterintuitive multimodal data, we face a core challenge: data directly collected from natural scenes typically follow statistical co-occurrence patterns, making it difficult to obtain representative counterintuitive samples. This collected approach makes it challenging to clearly distinguish whether the performance of visual language models (VLMs) in downstream tasks stems from a true understanding of image content or merely relies on statistical co-occurrence information for inference. To reduce the interference of common-sense bias, we employ a text-to-image generation approach to construct multimodal counterintuitive data, thereby controlling image content while breaking the statistical co-occurrence structure. Fig.~\ref{fig: pipeline} demonstrates the counterintuitive data collection process. In this section, we will provide a detailed overview of the ContraSense Benchmark construction.

\noindent \textbf{Prompt guides LLM to generate counterintuitive text.} To generate text with counterintuitive characteristics for the next stage of text-to-image generation tasks, we carefully designed a set of prompts to guide large language models in generating text that meets the requirements of the instructions. In constructing the prompts, we systematically constrained and designed the object categories, caption fields, structured labels, and six types of question-answering tasks for describing images in the captions. 

In the process of generating image captions, we guide large language models to generate image captions with counterintuitive characteristics by embedding instructions in the prompt. Specifically, as shown in Tab.~\ref{tab:counterintutive_categories}, we designed five categories of counterintuitive scenarios, including context violation (\textit{e.g.}, gorillas living underwater), attribute anomalies (\textit{e.g.}, naturally occurring purple lions), size ratio anomalies (\textit{e.g.}, a cat larger than a house), physically impossible scenarios (\textit{e.g.}, a floating piano), and functional state violations (\textit{e.g.}, ice burning). In the generation task, we randomly select one or more combinations of these counterintuitive categories to guide the large language model in fabricating a non-existing image and generating a caption that aligns with the selected counterintuitive settings.

In terms of object category settings, we require that each image contain at least two categories to support subsequent category relationship construction tasks. However, directly instructing a large language model to generate two common categories randomly results in a large number of duplicate categories in the generated output, exhibiting pseudo-random characteristics, thereby limiting category diversity. To enhance category generation diversity, we predefined 240 candidate categories and set the generation strategy in the prompt: one category is randomly generated by the LLM, while the other is randomly sampled from the predefined 240 categories with equal probability.

In terms of structured labels, we require the LLM to generate structured labels of different granularity for the two objects mentioned in the caption, including the object category, color, quantity, and a binary label indicating whether each object has counterintuitive characteristics. In addition, we describe the spatial relationship between the two objects in the overall scene to express the relative spatial information between objects more comprehensively.

\begin{table*}[t]
\begin{center}
\resizebox{\textwidth}{!}{
\begin{tabular}{c|l|l|l|l} 
\toprule
\textbf{No.} & \textbf{Type Name} & \textbf{Example} & \textbf{Description} & \textbf{Common Sense Detection Capability} \\
\midrule
1 & Context Violation & "A small wooden boat inside a modern bathroom, photorealistic" & Objects appearing in unlikely backgrounds or scenes & Scene Common Knowledge \\

2 & Attribute Violation & "A black banana" & Unreasonable attributes such as colour, material, orientation, etc. & Attribute recognition \\

3 & Size/Proportion Anomalies & "A giant cat towering over city buildings" & Proportion Extreme Anomalies & Scale Comparison \\

4 & Physics-Impossible & "A full glass of water floating in mid-air" & Violation of the laws of physics & Physics common sense \\

5 & Function/State Perversity & "A burning ice cube on a metal tray" & State does not correspond to the function of the object & State Identification \\

\bottomrule
\end{tabular}
}
\end{center}
\caption{ContraSense defines five counterintuitive categories, and each generated sample contains at least one of these categories, or a combination of multiple counterintuitive categories.}
\vspace{-0.3cm}
\label{tab:counterintutive_categories}
\end{table*}

In terms of designing counterintuitive questions, we embedded explicit instructions in the prompt. We set up six types of single-choice questions for each image, including object category, color attribute, quantity perception, whether the object is counterintuitive, and the spatial relationship between objects. In addition, we specially designed a higher-order question, namely the counterintuitive image-text matching question, aimed at directly assessing whether the model can accurately identify the most prominent counterintuitive features in the image. Each question provides five possible answers, with only one being the correct option. The purpose of this question design is to comprehensively evaluate whether the model truly answers based on the image content itself, rather than simply relying on superficial statistical co-occurrence relationships, across multiple dimensions—including category recognition, fine-grained attribute perception, quantity inference, nuanced understanding of multi-object relationships, added spatial reasoning, and the core counter-intuitive image–text matching.

\noindent \textbf{Text-to-image.} After using the above prompt to guide the large language model in generating text, we select the caption field and further employ a text-to-image generation method to generate high-definition images consistent with the semantic description of each given caption. Specifically, we input the caption into the generation model and set the image resolution to $1024 \times 1024$ pixels to ensure the output images have high clarity and rich detail. The model can comprehensively process multi-dimensional information such as objects, scenes, attributes, and spatial relationships mentioned in the caption, thereby generating images consistent with the caption's semantic meaning, ensuring the accuracy and reasonableness of the generated content. Using the above method, we ultimately generated 18,000 image-caption candidate pairs, which were then subjected to subsequent manual screening.

\noindent \textbf{Human calibration.} Existing large language models and text-to-image models have demonstrated remarkable text-generated expression capabilities and text-image matching capabilities. However, directly utilizing text generated by LLMs and images generated by text-to-image models as a counterintuitive dataset is not feasible, primarily because these generated results still contain significant biases. Specifically, during the text generation process of LLMs, the generated multiple-choice questions often exhibit issues such as text answer leakage, multiple answers in single-choice questions, and ambiguous referents. For example, the question \textit{What color is the green tiger in the image?} can be directly answered without even viewing the image. In text-to-image models, the most common issue is that the generated images do not fully match the given captions. To effectively address these issues, we have introduced a human-calibrated process. Specifically, we invited five human experts to independently review whether the generated images and text matched and conducted cross-validation to ensure consistency and accuracy between images and text. During this process, we removed samples where generation failed or where text and images did not match. Additionally, we corrected the text to ensure the accuracy of labels and the uniqueness of answers in multiple-choice questions. At the same time, to eliminate other potential confounding factors—such as the small scale of objects related to the question answers or their occupancy of only a small portion of the visual field—we excluded images where the answer-related content did not belong to the primary semantic object or was unrelated to it. Through human calibration, we ensured that all questions are directly related to the image content, especially those counterintuitive points that should be clear and explicit. Ultimately, we cleaned up 4,698 high-quality QA samples from the generated 18,000 image-caption pairs, forming our ContraSense Benchmark.

\section{Experiments}
This section provides a detailed description of our specific experimental setup and results for performance evaluation using multiple multimodal large models on the ContraSense benchmark.

\subsection{Experimental Settings and Ethical Statement}

\textbf{Evaluation Metrics}

\textbf{Standard Classification Metrics.} To comprehensively evaluate model performance under multi-class and class-imbalanced settings, we report a diverse set of metrics, including Accuracy, F1 Macro, F1 Weighted~\cite{sokolova2009systematic}, Precision, Recall, and Matthews Correlation Coefficient (MCC)~\cite{matthews1975comparison, gorodkin2004comparing}. These metrics capture complementary aspects of classification behavior, ranging from overall correctness to per-class balance and correlation-aware robustness.

\textbf{ROUGE~\cite{lin2005recall} \& METEOR~\cite{lavie2009meteor}.} To quantitatively evaluate the performance of VLM-generated text, we adopted standardized metrics commonly used in the field of natural language generation. Specifically, we used ROUGE-1, ROUGE-2, and ROUGE-L to measure the overlap between the generated text and the reference text at the unigram, bigram, and longest common subsequence levels. These metrics comprehensively reflect the similarity and fluency of the generated content. Additionally, we introduced the METEOR metric, which comprehensively evaluates semantic similarity through synonym matching and stemming, thereby providing a more detailed assessment of the content fidelity, information richness, and language quality of the generated text.

\textbf{Contra-Acc.} In image understanding tasks, models exhibit significant variations in performance across different task dimensions, reflecting their distinct capabilities when addressing diverse problem types. In multi-task scenarios, directly calculating average performance may conceal these shortcomings.  To comprehensively evaluate a model’s multidimensional image understanding capabilities and address the shortcoming of conventional evaluation metrics, which do not sufficiently account for the lower bounds of a model’s capabilities, we propose the Contra-Acc metric. We define the six-dimensional correctness vector for each sample as $\textbf{p}_i = (p_{i1}, p_{i2}, …, p_{i6}) \in [0, 1]^6$, and define the error vector as $\textbf{e}_i = \textbf{1} - \textbf{p}_i$. Based on this vector, we formulate the metric as follows:

\[
        S_i^\infty = \textbf{1} - \|\textbf{e}_i\|_\infty=\textbf{1} - \max _{1\leq j\leq 6}e_{ij} = \min_{1\leq j\leq 6}p_{ij}; \quad \frac{1}{N}\sum_{i=1}^N  S_i^\infty.
\]

We introduce a Chebyshev consistency metric based on the $L^\infty$ norm, leveraging the $\textbf{1}$-Lipschitz continuous kernel to prioritize the weakest dimension in modeling multidimensional consistency. Maximum deviation is employed to characterize consistency disparities across these dimensions. Contra-Acc offers a straightforward yet effective approach to pinpointing a model's weakest links, thereby enhancing our understanding and improving its multidimensional capabilities.

\textbf{Implemental details.} In terms of implementation details, we selected GPT-4.1~\cite{gpt4one} as the text generation large language model during the text generation stage to generate counterintuitive text. During the text-to-image generation stage, we employed FLUX.1-dev~\cite{flux2024} as the text-to-image model to generate high-resolution images that match the given caption. In the answer option extraction component of the text generation model, we detect the keywords A/B/C/D/E in the responses and ensure that no English letters appear before or after these keywords, using this as the criterion for option extraction. When no answer options are detected, option A is returned by default as the model's predicted result. This method has been previously applied in the literature~\cite{zhangvisually}. We used 1 NVIDIA L40s graphics card in the testing phase.

\begin{figure}[t]
    \centering
        \includegraphics[width=0.6\linewidth]{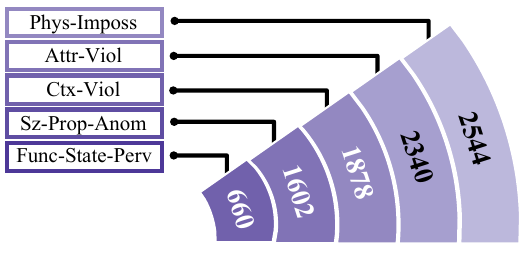}
    \vspace{-0.1cm}
    \caption{Distribution of different types of common sense violations in ContraSense. Phys-Imposs: Physics-Impossible. Attr-Viol: Attribute Violation. Ctx-Viol: Context Violation. Sz-Prop-Anom: Size/Proportion Anomalies. Func-State-Perv: Function/State Perversity.}
    \label{fig: contra-intutive data distribution}
\end{figure}

\subsection{Contra-intuitive Type Data Distribution}
Through our carefully designed counterintuitive data collection process, we have built the ContraSense dataset, which contains 4,698 high-quality QA pairs. Unlike large-scale datasets that focus on the sheer volume of images, ContraSense does not prioritize the size of the dataset but rather the richness, accuracy, and structured semantic annotations provided for each image. Specifically, each image undergoes multiple rounds of rigorous manual screening, modification, and detailed annotation. The annotation content covers six dimensions: image caption, the categories of two core objects, color, quantity, whether it violates common-sense attributes, and the semantic relationship between the two. Additionally, to thoroughly assess the model's understanding of image semantics, we designed six detailed multiple-choice questions for each image, comprehensively testing model performance across six dimensions: category, color, quantity, counterintuitive perception, object relationship understanding, and overall counterintuitive recognition.

To get a better understanding of how counterintuitive phenomena are spread out in the dataset, we manually labeled each image with its counterintuitive type. We counted how often each type showed up. As shown in Fig.~\ref{fig: contra-intutive data distribution}, since a single image can have more than one counterintuitive type, the total number of types is more than the total number of images (4,698), which shows that multiple types of counterintuitive phenomena can exist at the same time.

\begin{table*}[!t]
\resizebox{\textwidth}{!}{
\begin{tabular}{cccp{2.8cm}cccccc}
\toprule 
\hline
\multirow{2}{*}{\centering Model} & \centering \multirow{2}{*}{Image Encoder} & \centering \multirow{2}{*}{ Language Model} & \centering \multirow{2}{*}{Data Type} & \multicolumn{6}{c}{\textbf{ContraSense Bench} Performance} \\
\cline{5-10}
  & & & & Contra-Acc($\%$) $\uparrow$ & ROUGE-1 $\uparrow$ & ROUGE-2 $\uparrow$ & ROUGE-L $\uparrow$ & METEOR $\uparrow$ & Times\\
  
 \hline
 \centering Human Performance & - & - & \centering ContraSense & 96.04 & - & - & - & - & -\\
 \hline

 & & \centering \textbf{Open-source Models} & & & & & & & \\
\hline

  & & & \centering ContraSense              & 36.91 & 14.50 & 4.17 & 11.20 & 21.00 & 01:28:16\\
  & & & \centering Normal                       & 79.18 & 17.48 & 7.52 & 13.75 & 25.97 & 01:22:05\\
  \multirow{-3}{*}{LLaVA-NeXT-8B} & \multirow{-3}{*}{CLIP-L-14} & \multirow{-3}{*}{Llama-3-8B-Instruct} & \centering $\Delta$                         & -42.27 & -2.98 & -3.35 & -2.55 & -4.97 & - \\

  \rowcolor{gray!20}
  & & & \centering ContraSense              & 35.89 & 21.80 & 6.16 & 16.40 & 24.99 & 01:13:09\\
  \rowcolor{gray!20}
  & & & \centering Normal                       & 80.97 & 24.04 & 9.22 & 18.29 & 30.03 & 01:10:11\\
  \rowcolor{gray!20}
  \multirow{-3}{*}{DeepSeek-VL-7B-Chat} & \multirow{-3}{*}{SigLIP-L $\&$ SAM-B} & \multirow{-3}{*}{DeepSeek-LLM-7B} & \centering $\Delta$       & -45.08 & -2.24 & -3.06 & -1.89 & -5.04 & - \\

  & & & \centering ContraSense              & 37.55 & 29.57 & 7.59 & 22.21 & 20.36 & 00:52:07\\
  & & & \centering Normal                       & 81.74 & 33.98 & 11.68 & 26.71 & 25.11 & 00:51:35\\
  \multirow{-3}{*}{Llama-3.2-11B-Vision} & \multirow{-3}{*}{ViT-H-14} &  \multirow{-3}{*}{Llama3.1-8B} & \centering $\Delta$                           & -44.19 & -4.41 & -4.09 & -4.5 & -4.75 & -\\

  \rowcolor{gray!20}
  & & & \centering ContraSense              & 54.02 & 15.47 & 4.52 & 11.65 & 20.15 & 01:57:11\\
  \rowcolor{gray!20}
  & & & \centering Normal                       & 91.95 & 17.11 & 6.74 & 13.08 & 23.11 & 01:56:30\\
  \rowcolor{gray!20}
  \multirow{-3}{*}{Qwen2.5-VL-7B-Instruct} & \multirow{-3}{*}{ViT(SwiGLU, RMSNorm)} & \multirow{-3}{*}{Qwen2.5 LLM} & \centering $\Delta$           & -37.93 & -1.46 & -2.22 & -1.43 & -2.96 & -\\
 \hline
 
 & & \centering \textbf{Close-source Models} & & & & & & &\\
 \hline
  & & & \centering ContraSense              & 51.98 & 13.72 & 4.50 & 10.47 & 19.97 & 04:55:35\\
  & & & \centering Normal                       & 85.44 & 16.51 & 6.91 & 12.81 & 24.20 & 04:40:23\\
  \multirow{-3}{*}{Gemini-1.5-pro} &  \multirow{-3}{*}{-} & \multirow{-3}{*}{-} & \centering $\Delta$             
                                                & -33.46 & -2.79 & -2.41 & -2.34 & -4.23 & -\\
  \rowcolor{gray!20}
  & & & \centering ContraSense              & 47.13 & 13.96 & 3.92 & 10.55 & 18.82 & 03:24:17\\
  \rowcolor{gray!20}
  & & & \centering Normal                       & 80.72 & 16.71 & 6.25 & 12.78 & 22.45 & 03:14:00\\
  \rowcolor{gray!20}
  \multirow{-3}{*}{Gemini-2.0-flash} & \multirow{-3}{*}{-} & \multirow{-3}{*}{-} & \centering $\Delta$                                                    & -33.59 & -2.75 & -2.33 & -2.23 & -3.63 & - \\
  & & & \centering ContraSense              & 52.11 & 12.35 & 3.63 & 9.48 & 18.33 & 07:46:48\\
  & & & \centering Normal                       & 85.44 & 14.17 & 5.41 & 10.99 & 21.55 & 06:38:13\\
  \multirow{-3}{*}{GPT-4o} & \multirow{-3}{*}{-} & \multirow{-3}{*}{GPT-4} & \centering $\Delta$  
                                                & -33.33 & -1.82 & -1.78 & -1.51 & -3.22 & -\\
  \rowcolor{gray!20}
  & & & \centering ContraSense              & 60.03 & 12.95 & 3.89 & 9.78 & 19.05 & 06:44:14\\
  \rowcolor{gray!20}
  & & & \centering Normal                       & 86.08 & 15.34 & 5.71 & 11.60 & 22.97 & 06:43:13\\
  \rowcolor{gray!20}
  \multirow{-3}{*}{GPT-4.1} & \multirow{-3}{*}{-} & \multirow{-3}{*}{GPT-4} & \centering $\Delta$                                                      & -26.05 & -2.39 & -1.82 & -1.82 & -3.92 & -\\
 \hline

\bottomrule
\end{tabular}

}
\vspace{0.1cm}
\caption{Model performance evaluation on the ContraSense Benchmark. We systematically evaluated the overall performance of eight state-of-the-art VLM models on the ContraSense dataset and normal data. The $\Delta$ row indicates the difference between the two datasets across each metric. The experimental results show that all models perform significantly lower on the ContraSense dataset compared to the normal data, indicating that these models still have limited capability in accurately reflecting image content.}
\label{tab: main results}
\end{table*}

\subsection{Main Results}
\label{sec: main_results}
To comprehensively assess the impact of statistical co-occurrence on model performance, we created normal data that conformed to statistical co-occurrence characteristics and conducted systematic experiments on this data to compare differences with ContraSense data. Specifically, we followed the same prompt format as ContraSense and replaced the instructions for generating counterintuitive text with instructions for generating normal samples that conform to common sense. Since normal samples do not contain counterintuitive elements, we replaced counterintuitive questions with function questions. Other parts remained unchanged to minimize differences other than counterintuitive points. We use the same structured prompt template as ContraSense, while replacing the counter-intuitive constraints with commonsense-consistent instructions. The complete prompts are publicly available in the released benchmark repository. Subsequently, we conducted contrast experiments on the current state-of-the-art 8 VLM models using ContraSense and normal data. The experimental results are shown in Tab.~\ref{tab: main results}: On the Contra-Acc metric, open-source model performance on ContraSense data decreased by 37\%-45\% compared to normal data. Among these, LLaVA-NeXT-8B, DeepSeek-VL-7B-Chat, and Llama-3.2-11B-Vision achieved Contra-Acc scores of only 35\%-37\%, far below the average performance of approximately 80\% on normal data, failing to meet expectations. In contrast, closed-source models showed overall performance improvements, but Contra-Acc still decreased by approximately 30\% on ContraSense data. These results indicate that when data exceeds the scope of statistical co-occurrence, VLMs exhibit significant deficiencies in understanding and reasoning, often failing to remain faithful to the image content itself and being influenced by statistical co-occurrence bias. Comparison with Human Performance. They ultimately achieved an average Contra-Acc score of 96.04\% on the ContraSense benchmark. This result highlights the significant gap between the current VLM and humans in accurately reflecting image content. Such gaps could have catastrophic consequences in critical applications such as autonomous driving and medical image analysis, further emphasizing the necessity of enhancing the model's true visual understanding capabilities.\newline

\begin{table*}[!t]
\resizebox{\textwidth}{!}{
\begin{tabular}{cccp{4.0cm}ccccc}
\toprule 
\hline
\multirow{2}{*}{\centering Model} & \centering \multirow{2}{*}{Image Encoder} & \centering \multirow{2}{*}{ Language Model} & \centering \multirow{2}{*}{Data Type} & \multicolumn{5}{c}{\textbf{ContraSense Bench} Performance} \\
\cline{5-9}
  & & & & Contra-Acc($\%$) $\uparrow$ & ROUGE-1 $\uparrow$ & ROUGE-2 $\uparrow$ & ROUGE-L $\uparrow$ & METEOR $\uparrow$\\
  
 \hline

 & & \centering \textbf{Open-source Models} & & & & & & \\
\hline

  & & & \centering Normal                         & 79.18 & 17.48 & 7.52 & 13.75 & 25.97 \\
  & & & \centering Context Violation              & 42.49 & 15.08 & 4.34 & 11.40 & 21.42 \\
  & & & \centering Attribute Violation            & 38.10 & 15.00 & 4.36 & 11.50 & 21.51 \\
  & & & \centering Size/Proportion Anomalies      & 34.83 & 14.35 & 4.23 & 11.22 & 20.89 \\
  & & & \centering Physics-Impossible             & 35.61 & 14.22 & 3.98 & 11.06 & 20.66 \\
  \multirow{-6}{*}{LLaVA-NeXT-8B} & \multirow{-6}{*}{CLIP-L-14} & \multirow{-6}{*}{Llama-3-8B-Instruct} & \centering Function/State Perversity 
                                                  & 32.73 & 13.72 & 3.79 & 10.63 & 20.65 \\

  \rowcolor{gray!20}
  & & & \centering Normal                         & 80.97 & 24.04 & 9.22 & 18.29 & 30.03 \\
  \rowcolor{gray!20}
  & & & \centering Context Violation              & 41.85 & 22.32 & 6.45 & 16.45 & 25.27 \\
  \rowcolor{gray!20}
  & & & \centering Attribute Violation            & 33.58 & 22.61 & 6.68 & 16.83 & 25.71 \\
  \rowcolor{gray!20}
  & & & \centering Size/Proportion Anomalies      & 32.21 & 21.72 & 6.02 & 16.35 & 24.58 \\  \rowcolor{gray!20}
  & & & \centering Physics-Impossible             & 39.15 & 21.18 & 5.68 & 15.99 & 24.26 \\
  \rowcolor{gray!20}
  \multirow{-6}{*}{DeepSeek-VL-7B-Chat} & \multirow{-6}{*}{SigLIP-L $\&$ SAM-B} & \multirow{-6}{*}{DeepSeek-LLM-7B} & \centering Function/State Perversity                                                                                                            & 32.73 & 20.46 & 5.19 & 15.64 & 24.28 \\

  & & & \centering Normal                         & 81.74 & 33.98 & 11.68 & 26.71 & 25.11 \\
  & & & \centering Context Violation              & 40.58 & 29.94 & 7.68 & 22.35 & 20.15 \\
  & & & \centering Attribute Violation            & 34.59 & 29.83 & 7.84 & 22.23 & 20.29 \\
  & & & \centering Size/Proportion Anomalies      & 33.33 & 28.99 & 7.58 & 21.98 & 19.71 \\
  & & & \centering Physics-Impossible             & 39.15 & 29.17 & 7.55 & 21.98 & 20.48 \\
  \multirow{-6}{*}{Llama-3.2-11B-Vision} & \multirow{-6}{*}{ViT-H-14} &  \multirow{-6}{*}{Llama3.1-8B} & \centering Function/State Perversity      
                                                  & 38.18 & 29.17 & 7.15 & 22.33 & 21.18 \\

  \rowcolor{gray!20}
  & & & \centering Normal                         & 91.95 & 17.11 & 6.74 & 13.08 & 23.11 \\
  \rowcolor{gray!20}
  & & & \centering Context Violation              & 57.51 & 15.95 & 4.67 & 11.82 & 20.65 \\
  \rowcolor{gray!20}
  & & & \centering Attribute Violation            & 52.13 & 15.94 & 4.69 & 11.89 & 20.48 \\
  \rowcolor{gray!20}
  & & & \centering Size/Proportion Anomalies      & 50.56 & 15.18 & 4.40 & 11.54 & 19.76 \\
  \rowcolor{gray!20}
  & & & \centering Physics-Impossible             & 54.72 & 15.17 & 4.38 & 11.38 & 19.89 \\
  \rowcolor{gray!20}
  \multirow{-6}{*}{Qwen2.5-VL-7B-Instruct} & \multirow{-6}{*}{ViT(SwiGLU, RMSNorm)} & \multirow{-6}{*}{Qwen2.5 LLM} & \centering Function/State Perversity
                                                  & 50.00 & 15.18 & 4.28 & 11.40 & 20.30 \\
 \hline
 
 & & \centering \textbf{Close-source Models} & & & & & &\\
 \hline
  & & & \centering Normal                         & 85.44 & 16.51 & 6.91 & 12.81 & 24.20 \\
  & & & \centering Context Violation              & 53.35 & 13.45 & 4.56 & 10.13 & 19.78 \\
  & & & \centering Attribute Violation            & 51.38 & 13.62 & 4.61 & 10.33 & 19.94 \\
  & & & \centering Size/Proportion Anomalies      & 52.06 & 13.76 & 4.45 & 10.59 & 19.88 \\
  & & & \centering Physics-Impossible             & 51.18 & 13.94 & 4.41 & 10.64 & 20.26 \\
  \multirow{-6}{*}{Gemini-1.5-pro} &  \multirow{-6}{*}{-} & \multirow{-6}{*}{-} & \centering Function/State Perversity                                      & 46.36 & 13.28 & 4.13 & 10.16 & 19.24 \\
  \rowcolor{gray!20}
  & & & \centering Normal                         & 80.72 & 16.71 & 6.25 & 12.78 & 22.45 \\
  \rowcolor{gray!20}
  & & & \centering Context Violation              & 49.20 & 13.96 & 4.17 & 10.45 & 19.04 \\
  \rowcolor{gray!20}
  & & & \centering Attribute Violation            & 43.86 & 14.26 & 4.00 & 10.71 & 19.32 \\
  \rowcolor{gray!20}
  & & & \centering Size/Proportion Anomalies      & 47.19 & 13.60 & 3.95 & 10.47 & 18.42 \\
  \rowcolor{gray!20}
  & & & \centering Physics-Impossible             & 47.41 & 13.97 & 3.80 & 10.57 & 18.66 \\
  \rowcolor{gray!20}
  \multirow{-6}{*}{Gemini-2.0-flash} & \multirow{-6}{*}{-} & \multirow{-6}{*}{-} & \centering Function/State Perversity                       & 40.00 & 13.30 & 3.46 & 10.02 & 18.07 \\
  & & & \centering Normal                         & 85.44 & 14.17 & 5.41 & 10.99 & 21.55 \\
  & & & \centering Context Violation              & 55.27 & 12.33 & 3.64 & 9.34 & 18.51 \\
  & & & \centering Attribute Violation            & 52.13 & 12.25 & 3.64 & 9.41 & 18.43 \\
  & & & \centering Size/Proportion Anomalies      & 52.06 & 12.33 & 3.57 & 9.46 & 18.13 \\
  & & & \centering Physics-Impossible             & 50.71 & 12.62 & 3.67 & 9.72 & 18.39 \\
  \multirow{-6}{*}{GPT-4o} & \multirow{-6}{*}{-} & \multirow{-6}{*}{GPT-4} & \centering Function/State Perversity                                      & 40.00 & 12.18 & 3.26 & 9.27 & 18.05 \\
  \rowcolor{gray!20}
  & & & \centering Normal                         & 86.08 & 15.34 & 5.71 & 11.60 & 22.97 \\
  \rowcolor{gray!20}
  & & & \centering Context Violation              & 62.30 & 12.77 & 3.86 & 9.60 & 18.91 \\
  \rowcolor{gray!20}
  & & & \centering Attribute Violation            & 58.90 & 12.82 & 3.89 & 9.70 & 18.98 \\
  \rowcolor{gray!20}
  & & & \centering Size/Proportion Anomalies      & 62.17 & 13.00 & 3.92 & 9.90 & 19.12 \\
  \rowcolor{gray!20}
  & & & \centering Physics-Impossible             & 61.56 & 13.23 & 3.90 & 9.96 & 19.25 \\
  \rowcolor{gray!20}  
  \multirow{-6}{*}{GPT-4.1} & \multirow{-6}{*}{-} & \multirow{-6}{*}{GPT-4} & \centering Function/State Perversity                                      & 45.45 & 12.37 & 3.48 & 9.34 & 18.24 \\
 \hline

\bottomrule
\end{tabular}

}
\caption{Model performance was evaluated on the ContraSense dataset for different types of counterintuitive samples.}
\vspace{-0.1cm}
\label{tab: six violation acc}
\end{table*}

\textbf{Model Performance Across Violation Types.}
As for the caption generation evaluation, the performance of the four text matching metrics, ROUGE-1, ROUGE-2, ROUGE-L, and METEOR, on the ContraSense dataset is generally significantly degraded compared to the normal data. Specifically, the open-source models of ROUGE-1, ROUGE-2, ROUGE-L, and METEOR show an average performance degradation ranging from 1.5\% to 5\%. The closed-source models also show a similar downward trend of about 1.5\% to 4.2\%. This overall decline suggests that the models face challenges not only in choosing answers but also in the quality of the generated text when faced with the counterintuitive graphic matching task. Especially in the ContraSense, literal overlap and semantic alignment between the model-generated answers and the reference answers are disturbed by statistical co-occurrence bias, which makes it challenging to reflect the real information of the images faithfully. This  result further reveals the general limitations of current VLMs in dealing with counterintuitive scenarios.

To more clearly analyze the specific impact of the five types of contra-intuitive defined in the ContraSense dataset on the model inference process, we conducted performance evaluations for each type of anomaly separately.  As shown in Tab.~\ref{tab: six violation acc}, compared to normal samples, data from all contra-intuitive types resulted in varying degrees of performance degradation in the model. This finding indicates that current multimodal models still exhibit significant limitations in reasoning capabilities when faced with semantically contra-intuitive or counter-intuitive scenarios, particularly in terms of semantic robustness and common-sense consistency, where there is considerable room for improvement.

Context violation represents a relatively easier type of counterintuitive case to identify. Most models—including open-source ones such as LLaVA-NeXT and Qwen2.5-VL, as well as closed-source models like GPT-4.1 and Gemini—achieve relatively high accuracy on this category. We speculate that this is because semantic conflicts between objects and scenes are typically more explicit, making it easier for models to localize these counterintuitive regions and perform semantic reasoning. Even in the absence of explicit training for counterintuitive reasoning, models may still detect semantic violations by leveraging statistical co-occurrence patterns learned during pretraining.
Attribute violations and size/proportion anomalies consistently yield lower performance, particularly in open-source models. These two types of anomalies often involve fine-grained perceptual cues (\textit{e.g.}, abnormal colors), requiring detailed visual discrimination and precise cross-modal alignment. Their subtle nature further increases the difficulty of detection.

Recognition performance on the Physics-Impossible category is moderate—slightly better than on attribute-related anomalies, yet lower than on context violations. These anomalies, such as objects floating in mid-air or defying gravity, are often visually salient; however, the model's ability to interpret their implausibility depends on whether it encodes physical common sense, which is typically not explicitly represented during pre-training.

Function/State Perversity poses the greatest challenge, yielding the lowest accuracy across all models. These anomalies involve atypical combinations of object functionality and state logic (\textit{e.g.}, using a banana as a phone), often requiring reasoning that goes beyond superficial visual cues and depends on action understanding and commonsense knowledge. Most models exhibit near-random performance on such samples, reflecting fundamental limitations in semantic reasoning capabilities.

Overall, while closed-source models (\textit{e.g.}, GPT-4.1 and the Gemini series) consistently outperform open-source counterparts across all anomaly types, they still face significant bottlenecks in handling Function/State Perversity and attribute-based anomalies. This suggests that although larger model scales and pre-training data improve robustness to some extent, current models continue to struggle with generalization in semantically complex, fine-grained, and commonsense-reliant scenarios.

\begin{figure*}[t]
  \centering

   \includegraphics[width=0.7\linewidth]{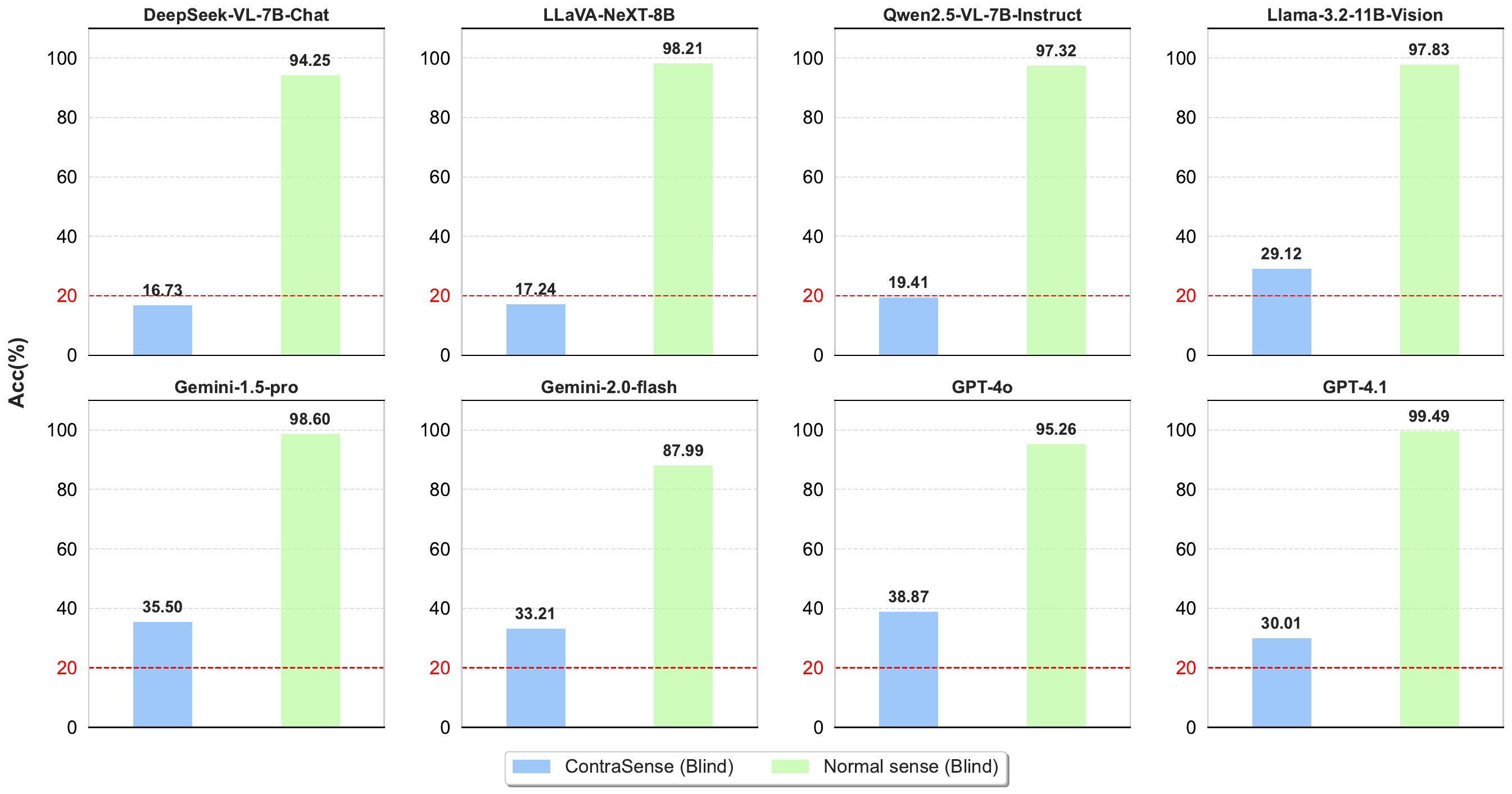}
    \vspace{-0.3cm}
   \caption{We replaced the input images with noisy images and conducted blind tests on ANOMALY RECOGNITION QUESTIONS and COMMONSENSE REASONING QUESTION. The results showed that the model's blind test performance on the normal data was significantly higher than its performance on the ContraSense dataset, averaging a high level of 95\%, indicating that even in the absence of image content, the model can still obtain the correct answer based on statistical co-occurrence. The red dashed line represents the randomly selected probability baseline.}
   \label{fig: blind}
\end{figure*}

\subsection{Vision-free Testing: Exposing Statistical Co-occurrence Reliance}
\textbf{Image Ablation.} In order to deeply understand the impact of statistical co-occurrence on model performance, we designed and implemented image ablation experiments. In terms of the experimental setup, for image input, we replaced all images in the ContraSense and normal data with noise images, completely dissociating them from the text content to ensure that no information relevant to answering the question could be obtained from the images. In terms of question selection, the five categories of questions—Object Class, Object Colour, Object Num, Is Counterintutive, and Relation — focus on examining image content from different dimensions. However, their answers do not necessarily involve counterintuitive points. Therefore, it is difficult to directly compare the differences in image content between normal questions and counterintuitive questions, making it insufficient to comprehensively validate the impact of statistical co-occurrence under conditions where image content is missing. For this purpose, we selected the Anomaly Recognition Question (corresponding to the Commonsense Reasoning Question in the normal dataset) as the comparison question. This question necessarily involves counterintuitive queries in the ContraSense group, which helps minimize the interference of common sense and thus more accurately assess the impact of statistical co-occurrence. In the experiment, we excluded the explicit influence of statistical co-occurrence in the ContraSense group from both image and text perspectives, using noise images that contain no image information for comparison testing. The experimental results are shown in Fig.~\ref{fig: blind}. On the ContraSense dataset, when image content is unavailable, the model's performance is significantly lower than that on normal data, indicating that the model cannot infer the answers to the Anomaly Recognition Question in ContraSense solely based on statistical co-occurrence. Remarkably, on the normal data, even when image content is unavailable, the model can still achieve an average performance exceeding 95\% by relying on statistical co-occurrence. \textbf{This result indicates that statistical co-occurrence can significantly enhance model performance under conditions where image information is unavailable, enabling the model to make correct judgments in specific tasks without relying on real visual information.}

\begin{figure*}[t]
  \centering

   \includegraphics[width=0.75\linewidth]{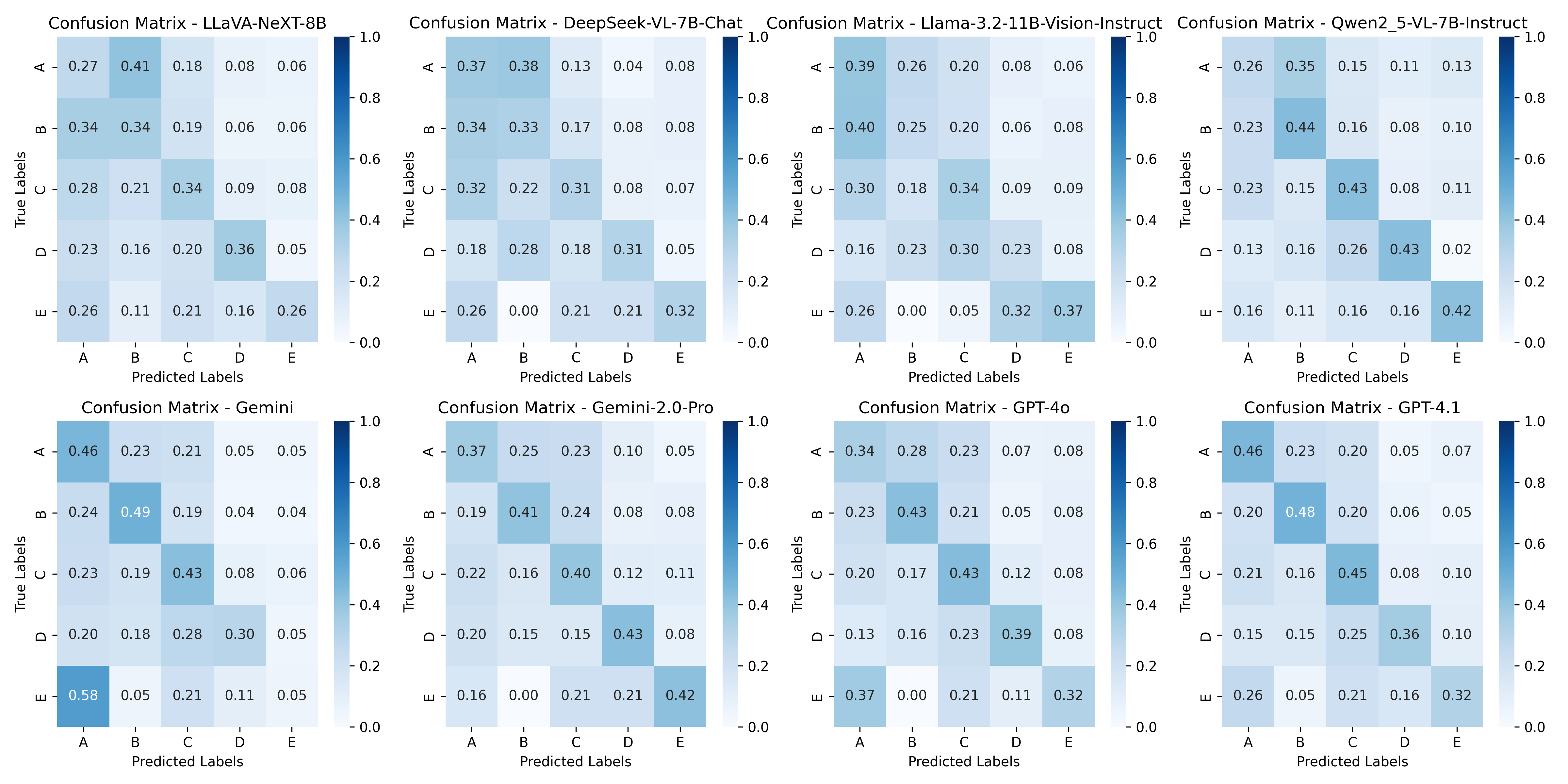}
    \vspace{-0.3cm}
   \caption{Row-normalized confusion matrices for eight MLLMs on the ContraSense benchmark. Each row corresponds to the true answer slot (A–E), and each column denotes the predicted answer slot.}
   \label{fig: confusion_matrix}
\end{figure*}

\subsection{Analysis of Performance Failures}
In Section~\ref{sec: main_results}, we analyzed the phenomenon of widespread systematic performance degradation when MLLMs encounter problems that significantly violate the statistical co-occurrence-based knowledge prior. Furthermore, the marked differences in performance range across different problem types indicate a highly negative correlation between the degree of deviation from statistical co-occurrence structures and model performance on each problem type: the more severe the deviation, the more pronounced the decline in model performance. The root cause lies in the model's inability to establish stable, generalizable mappings when semantic statistical co-occurrence is disrupted—whether between objects within an image or between images and text—leading to performance degradation. However, in addition to the aforementioned visual-text or object-object level semantic statistical co-occurrence, there exists another easily overlooked yet equally critical form of statistical co-occurrence: statistical co-occurrence at the level of answer choice positions. This co-occurrence does not originate from visual or semantic content itself, but manifests as systematic biases toward specific answer choice positions during the model's multi-choice decision-making process. 

In addition to semantic co-occurrence between visual objects and textual concepts, multiple-choice models may also exploit statistical regularities associated with answer positions. We therefore analyze these two effects separately. A complementary Macro-F1 analysis, which reduces the influence of uneven predictions across answer slots, confirms that the performance ranking across question types remains broadly consistent with that obtained using Accuracy; complete per-model results and detailed discussion are provided in Section B and Table VI of the Supplementary Material. In the following, we focus on the second effect and examine systematic option-position preferences through row-normalized confusion matrices.

\textbf{Option-level statistical co-occurrence.} To analyze the impact of option-level statistical co-occurrence on model behavior at the decision-making level, we constructed confusion matrices for the outputs of 8 MLLMs. We visualized the results, as shown in Fig.~\ref{fig: confusion_matrix}. Where each row represents the option position corresponding to the true label, and each column represents the option position predicted by the model (predicted label). Since options A–E correspond to different candidate answers across questions and do not represent fixed semantic categories, we applied row normalization to the confusion matrix. This allows us to characterize the conditional dispersion and misalignment patterns of model predictions across remaining slot positions, given the known true label slot. From this conditional probability perspective, the confusion matrix is no longer used to analyze confusion between semantic categories. Instead, it diagnoses non-semantic systematic behavioral patterns in the model during multi-choice decision scenarios, such as position bias and slot sensitivity. Furthermore, Table \ref{tab:option_bias} reports statistical results obtained by column-wise aggregation of row-normalized confusion matrices. Since A–E do not correspond to fixed semantic classes, these column sums reflect the model's overall selection tendencies toward each option slot under different ground-truth conditions. They can be interpreted as relative preference intensity. Theoretically, if the model exhibits no option slot bias, each column sum should approach 1.

\begin{table}[t]
\resizebox{8.8cm}{!}{
\centering
\begin{tabular}{lccccc}
\toprule
\textbf{Model} & \textbf{A} & \textbf{B} & \textbf{C} & \textbf{D} & \textbf{E} \\
\midrule
Open-source & & & & & \\
\hline
LLaVA-NeXT-8B                & 1.381 & 1.231 & 1.127 & 0.746 & 0.515 \\
DeepSeek-VL-7B-Chat          & 1.478 & 1.207 & 1.000 & 0.713 & 0.603 \\
Llama-3.2-11B-Vision-Instruct& 1.523 & 0.919 & 1.092 & 0.781 & 0.685 \\
Qwen2.5-VL-7B-Instruct       & 1.016 & 1.200 & 1.151 & 0.859 & 0.774 \\
\hline
Closed-source & & & & & \\
\hline
Gemini                       & 1.716 & 1.143 & 1.317 & 0.570 & 0.254 \\
Gemini-2.0-Pro               & 1.129 & 0.964 & 1.229 & 0.930 & 0.748 \\
GPT-4o                       & 1.264 & 1.039 & 1.316 & 0.739 & 0.643 \\
GPT-4.1                      & 1.284 & 1.062 & 1.310 & 0.716 & 0.629 \\
\hline
\bottomrule
\end{tabular}
}
\caption{Column-wise sums of row-normalized confusion matrices, reflecting models' positional bias over options.}
\vspace{-0.6cm}
\label{tab:option_bias}
\end{table}

Experimental results show that most models exhibit a significant preference for options A and C, while consistently avoiding the last option. Column E ranks lowest or second-lowest across nearly all models. Empirically, this phenomenon may relate to the distribution of question formats in training data: existing large-scale training corpora predominantly feature four-choice questions (A–D). In contrast, multiple-choice questions with an E option are relatively scarce. This induces statistical co-occurrence of specific option positions at the decision layer during training. This indicates that current models remain influenced by non-semantic positional priors during multiple-choice decision-making, with this bias originating from the structure of the options themselves rather than failures in understanding visual or semantic content. Further comparison across model types reveals that closed-source models generally exhibit a concentrated preference for A/C options, with their consistent distribution across rows indicating higher stability of this bias. In contrast, open-source models exhibit mixed preferences for A/B/C options, with more dispersed bias distributions and more random error transfer. This comparison indicates that stronger models do not eliminate option position bias but instead solidify it in a more stable and consistent form.

In summary, through column-wise aggregation analysis of the confusion matrix, we observe a systematic bias toward certain option positions that is consistent across models: models generally exhibit significant attraction to the first and middle options while systematically avoiding the last option. When statistical co-occurrence at the semantic or visual level is disrupted, models often degrade into relying on the statistical frequency or linguistic prior of the answer slot position itself, and frequently fall back to fixed options. This creates a non-causal statistical co-occurrence structure within the decision space, ultimately leading to systematic positional bias.

\section{Conclusion}
In this paper, we introduce a brand-new benchmark dataset: the ContraSense Benchmark. By constructing counterintuitive scenarios under multi-object conditions, we have maximized the elimination of the influence of statistical co-occurrence on question answering, thereby enabling a more accurate assessment of the model's ability to understand image content faithfully. Additionally, we propose the Contra-Acc metric, designed to comprehensively quantify the model's understanding of image content across multiple dimensions. Experimental results demonstrate that even current state-of-the-art models significantly lag behind human performance in terms of image content understanding.

\bibliography{IEEEtrans}
\bibliographystyle{IEEEtran}

\clearpage

\clearpage

\begin{strip}
\begin{center}
{\Huge Can Vision–Language Models Truly Understand Images? A Benchmark Beyond Commonsense Bias\par}
\vspace{1.2em} 
{\LARGE Supplementary Material\par}
\vspace{1em}
{\large Ling Lin, Heng Su, Xiangzeng Liu, Yang Bai, Congcong Zhu, Yang Long, Senior Member, IEEE, \\
Xin Wang, Member IEEE, Ling Shao, Fellow IEEE, Huazhu Fu, Senior Member, IEEE, Jingrun Chen}
\end{center}
\end{strip}

\setcounter{section}{0}

\section{Comparison With Existing Multimodal Benchmarks}
ContraSense has significant advantages over existing benchmarks in three key areas. First, our benchmark supports multiple answer types, covering both multiple-choice and open-ended task formats. In contrast, most current multimodal evaluation benchmarks only provide open-ended annotations, making it challenging to explore differences in model behavior across different answer settings effectively. Diverse answer types enable more granular capability assessment. Second, ContraSense focuses on contra-intuitive reasoning tasks, a capability that is currently underrepresented in mainstream evaluation frameworks. Existing benchmarks primarily focus on conventional text-image question-answering or description generation tasks, lacking structured evaluation for content that violates common-sense semantics. Therefore, ContraSense fills an important gap in this field, enabling the evaluation of a model's understanding and reasoning capabilities when faced with semantic anomalies and counter-intuitive scenarios. This minimizes the influence of statistical co-occurrence on model performance. Finally, our dataset employs a fully manual annotation process, covering the construction of image-text pairs, the design of questions and answers, and the manual classification of anomaly types. The annotation process involves human annotators, ensuring semantic consistency and annotation quality, providing a solid foundation for reliable evaluation. Tab.~\ref{tab: comparison_other_bench} compares ContraSense with multiple mainstream MLLM benchmarks, highlighting our systematic advantages in semantic counter-intuitive coverage, multi-task support, and annotation accuracy.

\begin{table*}[b]
\centering
\resizebox{13.5cm}{!}{
\begin{tabular}{ccccc} 
\toprule 
Models & Data Scale & Answer Type & Contra-intuitive & Human Annotation \\
\midrule 
HumanEval~\cite{chen2021evaluating} & 164 & Open & \textcolor{red}{\ding{55}}  & \textcolor{red}{\ding{55}} \\
MM-Vet~\cite{yu2023mm} & 218 & Open & \textcolor{red}{\ding{55}} & \textcolor{red}{\ding{55}} \\
INTERLEAVEDBENCH~\cite{liu2024holistic} & 815 & Open & \textcolor{red}{\ding{55}} & \textcolor{green}{\ding{51}}\\
MMIE~\cite{xia2024mmie} & 20K & Multi-Choice \& Open & \textcolor{red}{\ding{55}} & \textcolor{green}{\ding{51}}\\
\hline
\textbf{ContraSense (Ours)} & 4,698 & Multi-Choice \& Open & \textcolor{green}{\ding{51}} & \textcolor{green}{\ding{51}}\\
\bottomrule
\end{tabular}
}
\caption{The comparison between ContraSense with other LVLM benchmarks.}
\vspace{-0.3cm}
\label{tab: comparison_other_bench}
\end{table*}

\begin{table*}[!t]
\resizebox{\textwidth}{!}{
\begin{tabular}{cccp{4.0cm}cccccc}
\toprule 
\hline
\multirow{2}{*}{\centering Model} & \centering \multirow{2}{*}{Image Encoder} & \centering \multirow{2}{*}{ Language Model} & \centering \multirow{2}{*}{Data Type} & \multicolumn{5}{c}{\textbf{ContraSense Bench} Performance} \\
\cline{5-10}
  & & & & Accuracy($\%$) $\uparrow$ & F1 Macro($\%$) $\uparrow$ & F1 Weighted($\%$) $\uparrow$ & Precision($\%$) $\uparrow$ & Recall($\%$) $\uparrow$ & MCC($\%$) $\uparrow$\\
  
 \hline

 & & \centering \textbf{Open-source Models} & & & & & & \\
\hline

  & & & \centering Object Class                   & 19.05 & 12.62 & 21.02 & 14.47 & 11.89 & -7.72 \\
  & & & \centering Object Colour                  & 30.00 & 25.16 & 32.17 & 27.47 & 28.72 & 11.01 \\
  & & & \centering Object Num                     & 37.50 & 33.01 & 41.34 & 38.70 & 30.03 & 18.36 \\
  & & & \centering Counterintuitive               & 36.05 & 27.53 & 36.47 & 27.38 & 27.91 & 9.06 \\
  & & & \centering Relation                       & 47.62 & 34.42 & 54.55 & 36.84 & 42.61 & 29.41 \\
  & & & \centering Anomaly Recognition            & 24.07 & 19.62 & 24.88 & 22.16 & 19.99 & -2.00 \\
  \multirow{-7}{*}{LLaVA-NeXT-8B} & \multirow{-7}{*}{CLIP-L-14} & \multirow{-7}{*}{Llama-3-8B-Instruct} & \centering Overall Performance                                     & 32.77 & 29.10 & 34.38 & 29.50 & 31.58 & 10.54 \\

  \rowcolor{gray!20}
  & & & \centering Object Class                   & 31.75 & 21.76 & 32.31 & 22.51 & 22.73 & 8.89 \\
  \rowcolor{gray!20}
  & & & \centering Object Colour                  & 29.00 & 25.14 & 32.21 & 28.24 & 29.05 & 10.14 \\
  \rowcolor{gray!20}
  & & & \centering Object Num                     & 52.08 & 44.39 & 54.47 & 46.27 & 57.43 & 38.43 \\
  \rowcolor{gray!20}
  & & & \centering Relation                       & 34.29 & 22.93 & 41.56 & 26.92 & 21.96 & 9.95 \\
  \rowcolor{gray!20}
  & & & \centering Counterintuitive               & 37.77 & 34.22 & 37.72 & 33.02 & 38.56 & 11.01 \\
  \rowcolor{gray!20}
  & & & \centering Anomaly Recognition            & 20.99 & 17.43 & 21.68 & 20.30 & 19.87 & -2.68 \\
  \rowcolor{gray!20}
  \multirow{-7}{*}{DeepSeek-VL-7B-Chat} & \multirow{-7}{*}{SigLIP-L $\&$ SAM-B} & \multirow{-7}{*}{DeepSeek-LLM-7B} & \centering Overall Performance                  & 32.63 & 29.31 & 33.99 & 30.26 & 32.66 & 11.27 \\

  & & & \centering Object Class                   & 34.92 & 23.45 & 37.04 & 26.36 & 23.86 & 13.29 \\
  & & & \centering Object Colour                  & 27.00 & 20.04 & 29.35 & 22.66 & 20.29 & 5.15 \\
  & & & \centering Object Num                     & 27.08 & 22.24 & 30.46 & 30.98 & 20.63 & 9.08 \\
  & & & \centering Counterintuitive               & 30.04 & 27.68 & 28.92 & 26.16 & 35.92 & 2.47 \\
  & & & \centering Relation                       & 39.05 & 26.88 & 46.33 & 31.12 & 33.41 & 17.38 \\
  & & & \centering Anomaly Recognition            & 32.72 & 30.25 & 34.49 & 31.74 & 37.82 & 13.36 \\
  \multirow{-7}{*}{Llama-3.2-11B-Vision} & \multirow{-7}{*}{ViT-H-14} &  \multirow{-7}{*}{Llama3.1-8B} & \centering Overall Performance                                     & 31.79 & 27.64 & 33.02 & 28.14 & 31.75 & 10.03 \\

  \rowcolor{gray!20}
  & & & \centering Object Class                   & 47.62 & 35.92 & 50.42 & 37.38 & 37.17 & 30.08 \\
  \rowcolor{gray!20}
  & & & \centering Object Colour                  & 46.00 & 36.39 & 48.91 & 38.66 & 39.17 & 29.29 \\
  \rowcolor{gray!20}
  & & & \centering Object Num                     & 56.25 & 46.56 & 57.85 & 51.58 & 48.79 & 42.73 \\
  \rowcolor{gray!20}
  & & & \centering Counterintuitive               & 35.62 & 30.81 & 36.31 & 29.22 & 46.94 & 11.16 \\
  \rowcolor{gray!20}
  & & & \centering Relation                       & 60.00 & 42.00 & 66.27 & 42.76 & 46.88 & 40.76 \\
  \rowcolor{gray!20}
  & & & \centering Anomaly Recognition            & 20.37 & 16.84 & 23.44 & 20.30 & 18.25 & -3.14 \\
  \rowcolor{gray!20}
  \multirow{-7}{*}{Qwen2.5-VL-7B-Instruct} & \multirow{-7}{*}{ViT(SwiGLU, RMSNorm)} & \multirow{-7}{*}{Qwen2.5 LLM} & \centering Overall Performance                  & 39.66 & 34.48 & 41.66 & 34.60 & 39.45 & 20.40 \\
 \hline
 
 & & \centering \textbf{Close-source Models} & & & & & &\\
 \hline
  & & & \centering Object Class                   & 42.86 & 29.99 & 45.66 & 31.56 & 29.00 & 21.57 \\
  & & & \centering Object Colour                  & 31.00 & 24.64 & 33.60 & 27.39 & 24.35 & 9.66 \\
  & & & \centering Object Num                     & 39.58 & 31.51 & 40.92 & 36.56 & 45.71 & 21.89 \\
  & & & \centering Counterintuitive               & 46.35 & 34.13 & 46.92 & 34.29 & 34.65 & 24.52 \\
  & & & \centering Relation                       & 30.48 & 21.99 & 34.33 & 24.32 & 32.72 & 7.53 \\
  & & & \centering Anomaly Recognition            & 55.56 & 46.87 & 56.18 & 48.50 & 49.82 & 40.49 \\
  \multirow{-7}{*}{Gemini-1.5-pro} &  \multirow{-7}{*}{-} & \multirow{-7}{*}{-} & \centering Overall Performance                                                              & 43.18 & 34.12 & 44.20 & 34.76 & 34.61 & 22.93 \\
  \rowcolor{gray!20}
  & & & \centering Object Class                   & 34.92 & 25.65 & 36.94 & 27.63 & 25.74 & 14.16 \\
  \rowcolor{gray!20}
  & & & \centering Object Colour                  & 33.00 & 30.36 & 35.06 & 32.13 & 35.38 & 16.11 \\  
  \rowcolor{gray!20}
  & & & \centering Object Num                     & 45.83 & 39.11 & 51.30 & 44.17 & 37.21 & 32.25 \\
  \rowcolor{gray!20}
  & & & \centering Counterintuitive               & 24.89 & 20.75 & 25.29 & 20.13 & 29.15 & -2.72 \\
\rowcolor{gray!20}
  & & & \centering Relation                       & 39.05 & 27.96 & 44.97 & 30.66 & 38.08 & 19.43 \\
  \rowcolor{gray!20}
  & & & \centering Anomaly Recognition            & 65.43 & 55.11 & 67.58 & 54.67 & 61.05 & 53.83 \\
  \rowcolor{gray!20}
  \multirow{-7}{*}{Gemini-2.0-flash} & \multirow{-7}{*}{-} & \multirow{-7}{*}{-} & \centering Overall Performance                                                             & 39.66 & 35.09 & 41.08 & 34.58 & 40.34 & 20.08 \\
  & & & \centering Object Class                   & 41.27 & 28.53 & 44.01 & 30.66 & 27.48 & 20.73 \\
  & & & \centering Object Colour                  & 32.00 & 25.87 & 34.68 & 27.85 & 25.68 & 12.25 \\
  & & & \centering Object Num                     & 35.42 & 29.63 & 39.31 & 33.78 & 28.35 & 17.71 \\
  & & & \centering Counterintuitive               & 28.33 & 22.32 & 28.50 & 22.11 & 31.43 & 2.68 \\
  & & & \centering Relation                       & 39.05 & 27.15 & 43.99 & 31.00 & 34.42 & 22.25 \\
  & & & \centering Anomaly Recognition            & 66.05 & 56.67 & 66.96 & 57.90 & 59.88 & 53.98 \\
  \multirow{-7}{*}{GPT-4o} & \multirow{-7}{*}{-} & \multirow{-7}{*}{GPT-4} & \centering Overall Performance                                                                   & 40.65 & 34.81 & 42.16 & 34.54 & 38.18 & 20.50 \\
  \rowcolor{gray!20}
  & & & \centering Object Class                   & 46.03 & 33.86 & 48.90 & 35.42 & 33.98 & 26.98 \\
  \rowcolor{gray!20}
  & & & \centering Object Colour                  & 35.00 & 27.55 & 38.11 & 29.91 & 26.61 & 15.22 \\
  \rowcolor{gray!20}
  & & & \centering Object Num                     & 41.67 & 34.88 & 47.83 & 45.10 & 32.03 & 27.34 \\
  \rowcolor{gray!20}
  & & & \centering Counterintuitive               & 30.90 & 23.71 & 31.84 & 23.95 & 24.87 & 4.70 \\
  \rowcolor{gray!20}
  & & & \centering Relation                       & 45.71 & 32.81 & 50.63 & 34.23 & 42.83 & 26.57 \\
  \rowcolor{gray!20}
  & & & \centering Anomaly Recognition            & 71.60 & 61.59 & 72.32 & 61.60 & 64.10 & 61.77 \\
  \rowcolor{gray!20}  
  \multirow{-7}{*}{GPT-4.1} & \multirow{-7}{*}{-} & \multirow{-7}{*}{GPT-4} & \centering Overall Performance                                                                  & 45.01 & 38.26 & 46.40 & 38.09 & 41.34 & 26.15 \\
 \hline

\bottomrule
\end{tabular}

}
\vspace{0.1cm}
\caption{Model performance was evaluated on the ContraSense dataset for different types of questions.}
\label{tab: diff type questions}
\end{table*}

\section{Per-Question Evaluation and Analysis}
\label{per_question_evaluation_and_analysis}
Contra-Acc is used to assess the lower bound of a model's capability across various task dimensions, thereby highlighting its weakest links. To further analyze the model's specific behavior across different problem types, we conducted independent performance tests for each of the six problem categories in the ContraSense benchmark: Object Class, Object Colour, Object Num, Counterintuitive, Relation, and Anomaly Recognition. The final row of the table shows the overall performance, representing the average results across all problem types. Since not all problems involve counterintuitive factors, we additionally manually selected a batch of image-problem pairs where the questions explicitly violate statistical co-occurrence patterns. These were used to evaluate the model's performance in such scenarios. The results are shown in Table~\ref{tab: diff type questions}, from the accuracy metric perspective, which measures the overall correct selection rate of the model, significant performance differences among various models can be observed across different question types. Even the best-performing model overall, GPT-4.1, achieved only 45.01\% accuracy; other models scored between 31\% and 43\%. This generally low performance indicates that current models exhibit significant degradation when confronted with inputs that deviate from statistical co-occurrence patterns. This suggests their predictive capabilities remain heavily reliant on semantic co-occurrence rather than genuine visual-semantic understanding.

Analysis based on the Accuracy metric indicates that, when comprehensively evaluating the performance of open-source and closed-source models across six problem types, their overall capabilities exhibit a descending order of performance: Anomaly Recognition $>$ Relation $\approx$ Object Num $>$ Counterintuitive $>$ Object Colour $>$ Object Class.

Across different problem types, nearly all models exhibit the lowest accuracy on Object Class tasks. Specifically, the accuracy scores of open-source models predominantly cluster within the 20\%–35\% range. While closed-source models generally outperform their open-source counterparts, their performance remains significantly inferior to that of other question types. This phenomenon suggests that when the task object category significantly deviates from statistical co-occurrence-based prior knowledge, current MLLMs exhibit systemic deficiencies in classifying counterintuitive objects.

Regarding the Object Colour task, despite color typically being considered a relatively "simple" perceptual attribute, overall accuracy remains low, with most models achieving around 30\%. This result suggests that such errors do not originate from failures in low-level perception, but are more likely related to erroneous binding between color and object, or prior interference introduced by anomalous color settings. Even for perceptual attribute tasks, model prediction accuracy drops significantly when encountering counterintuitive semantics or weakened object-attribute bindings.

For Counterintuitive problems, performance differences between models are relatively minor, with accuracy typically clustering between 30\% and 40\%. This suggests that models can perceive the presence of "counterintuitiveness" to some extent but struggle to make stable, generalizable, and correct decisions in such judgments.

For Object Num problems, nearly all models achieved significantly higher accuracy than Object Class, Object Colour, and Counterintuitive tasks. However, overall performance remains suboptimal, with only a few models achieving an accuracy of over 50\%. This indicates that current MLLMs still suffer widespread performance degradation when counting tasks involve counterintuitive objects.

For Relation problems, performance disparities among models widened further. Although these tasks do not yet involve complex spatial reasoning or multi-step logical inference, their difficulty extends beyond purely perceptual capabilities. The increased complexity amplified the differences in models' abilities.

In Anomaly Recognition tasks, nearly all models achieved the highest accuracy. This is closely tied to the nature of the task: such problems lean more toward holistic semantic judgment, typically presenting candidate descriptions in multiple-choice format and requiring models to match images with counterintuitive text given the image context.

Notably, the performance ranking across different problem types aligns closely with their degree of disruption to statistical co-occurrence structures. When problems directly introduce object category errors or counterintuitive color settings (\textit{e.g.}, Object Class and Object Colour), statistical co-occurrence is severely disrupted, leading to the lowest model performance. In contrast, Counterintuitive and Object Num problems, which involve anomalies only at the object attribute or behavioral level, yield somewhat improved performance. In contrast, Relation and Anomaly Recognition problems exhibit weaker disruption to statistical co-occurrence patterns, focusing more on semantic matching between images and descriptions, thus yielding relatively better performance. Overall, this trend indicates that current model predictions remain heavily reliant on statistical co-occurrence structures and have yet to develop a sufficient understanding of genuine visual semantics.

\textbf{Disentangles option position statistics from co-occurrence interference.} In Table~\ref{tab: diff type questions}, we report performance results based on F1 Macro. F1 Macro is defined as $F1_{\text{Macro}}=\frac{1}{N}\sum_{i=1}^N F1_i$, where $N$ denotes the number of categories. This metric employs an equal-weight average (also known as a non-weighted average) of the F1 scores across all categories (options). It is commonly used in traditional classification tasks to mitigate evaluation bias caused by class imbalance. It should be noted that in ContraSense's task design, options A–E do not correspond to fixed semantic categories (\textit{e.g.}, one option always representing cats or dogs). Instead, they denote discrete answer slots for the same question. Therefore, these options do not inherently represent high-frequency or long-tail categories in a semantic sense. Under this premise, F1 Macro no longer measures semantic category imbalance; instead, it imposes an equal-weight constraint on prediction behavior across different answer slots. Essentially, F1 Macro penalizes answer-position bias—the statistical dependency a model develops toward specific answer slots during multi-choice decision-making. Thus, F1 Macro can be viewed as a metric designed to suppress statistical co-occurrence at the answer slot level. By eliminating interference from statistical co-occurrence across slots, F1 Macro effectively removes decision-level statistical noise. This disentangles the statistical interference introduced by slot positions, making the impact of semantic co-occurrence deviation across different question types clearer and more distinguishable in terms of performance. Under this framework, across all models (including open-source and closed-source models) and six question types, the performance ranking based on F1 Macro exhibits high consistency with that of Accuracy. Compared to Accuracy, this ranking demonstrates greater stability while significantly amplifying performance differences between models. This phenomenon reinforces the conclusion drawn from Accuracy: current MLLMs' predictions remain heavily reliant on statistical co-occurrence structures, lacking sufficient understanding of genuine visual-semantic relationships. When question content deviates strongly from these statistical patterns, models inevitably experience systematic performance degradation.

\section{Impact of Resolution Variation} 
\label{impact_of_resolution_variation}
\begin{figure}[t]
  \centering
   \includegraphics[width=0.8\linewidth]{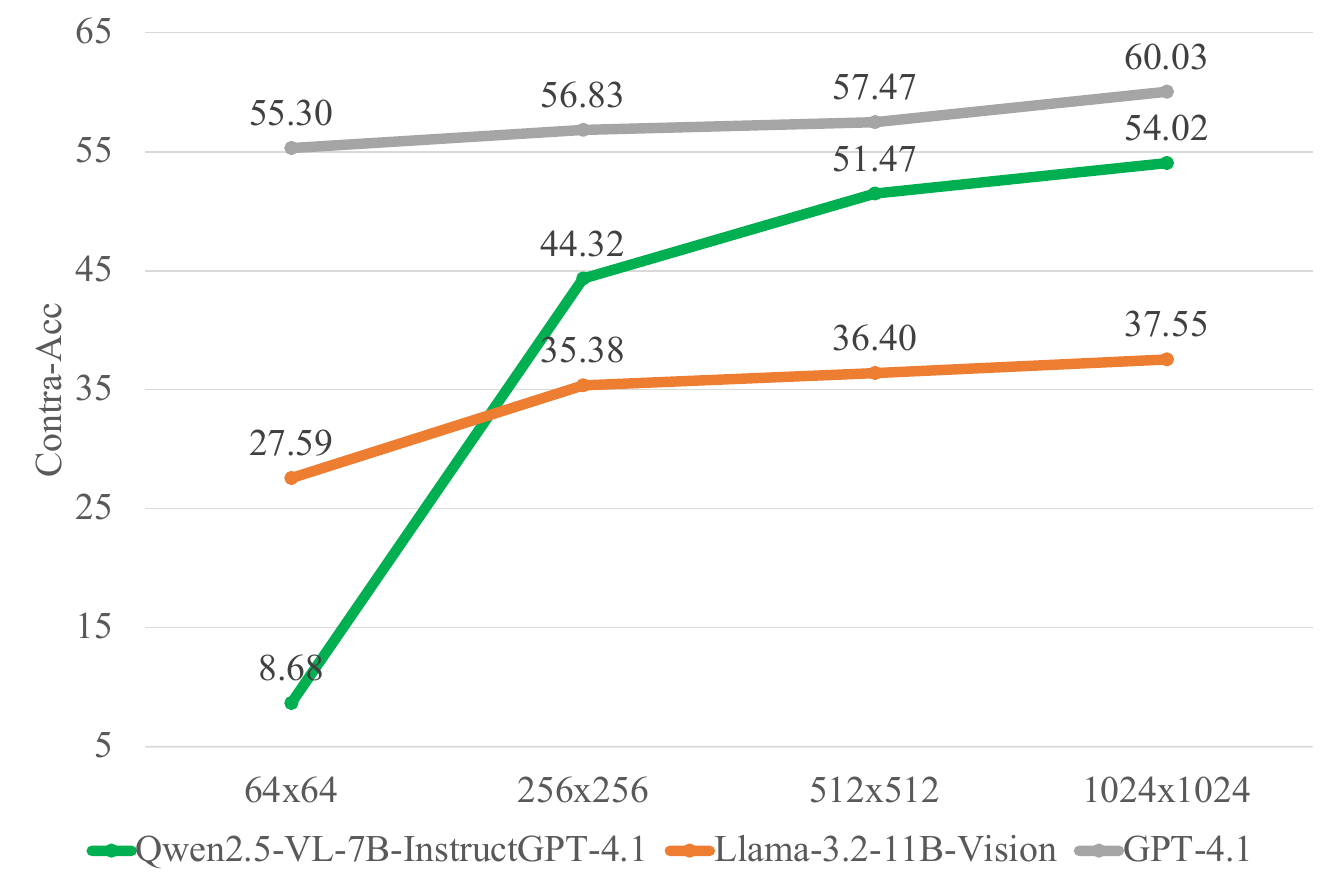}
    \vspace{-0.3cm}
   \caption{The figure shows the performance trends of mainstream multimodal models at different resolutions.
   }
   \label{fig: diff_resolution}
\end{figure}

To systematically assess the impact of image resolution on the performance of multimodal models, we conducted resolution ablation experiments on three types of models: Qwen2.5-VL-7B-Instruct, which uses a dynamic resolution strategy; LLaMA-3.2-11B-Vision, which uses a fixed resolution strategy; and the closed-source model GPT-4.1. We progressively reduced the input images to different resolutions and measured model performance on the ContraSense benchmark, with results shown in Fig.~\ref{fig: diff_resolution}.

From an overall trend perspective, all models exhibit varying degrees of performance degradation when the input resolution is reduced. This indicates that the models' perceptual capabilities are sensitive to the distribution of image details. Specifically, Qwen2.5-VL-7B-Instruct is the most sensitive to resolution changes, showing a significant decline in performance. When the input image resolution is reduced to $64 \times 64$, its Contra-Acc accuracy rate drops to just 8.68\%. This phenomenon highlights the limitations of dynamic resolution strategies when dealing with extremely low-resolution inputs: since their core objective is to balance perceptual capability with computational overhead, when images are forced to be compressed, local structures and micro-semantics are significantly lost, leading to overly sparse visual token representations. As a result, the model struggles to obtain sufficient semantic signals, ultimately causing a sharp decline in performance.

In contrast, LLaMA-3.2-11B-Vision employs a fixed-resolution input strategy, resulting in a more gradual decline in performance. This is primarily due to the fact that during training, the model consistently receives images of uniform dimensions (\textit{e.g.}, $224 \times 224$ or $448 \times 448$), enabling it to learn to extract stable and robust features at a specific scale. Even when the original resolution of the input images is reduced, after interpolation or scaling to a fixed size, it can still retain some key semantic structures, thereby maintaining relatively stable inference performance.

GPT-4.1's performance is least affected by resolution changes and remains almost stable. Empirical analysis suggests that this robustness may stem from its training process covering a wide distribution of image sizes, combined with diverse data augmentation strategies, enabling the model to achieve good scale generalization capabilities.

It is worth noting that ContraSense adopts $1024\times1024$ images, preserving micro-semantic differences (small objects, background cues, edge details) and producing denser patch-token representations. This design improves image-text alignment and enables high-resolution models such as GPT-4.1 to realize their potential on fine-grained image-text reasoning.

\section{Limitations}
\label{limitations}
The semi-automated nature of the ContraSense dataset construction process introduces considerations of scalability. While the introduction of manual screening and correction ensures high consistency between the pairing image and text and the clarity of counterintuitive content in the image, this approach demands significant human resource costs, hence the decrease in efficiency compared to a fully automated process. Moving forward, exploring more efficient construction workflows to reduce human resource consumption represents a valuable direction for future research.

\section{Ethical statement}
\label{Ethical_statement}
This research dataset is entirely generated by text-to-image models and contains no real individuals, geographic locations, or sensitive events. All content has undergone manual review and security filtering. All samples are used for academic research purposes to evaluate the model's semantic understanding and robustness, posing no public safety or value-oriented risks. Therefore, there are no potential ethical concerns.

\vfill

\end{document}